\documentclass[preprint,12pt]{elsarticle}

\usepackage{amssymb}
\usepackage{amsmath}
\usepackage{mathtools}
\usepackage{natbib}
\usepackage{url}
\usepackage[capitalise]{cleveref}
\usepackage{xcolor}
\usepackage{booktabs}
\usepackage{multirow}
\usepackage{multicol}
\usepackage{microtype}
\usepackage{algorithm}
\usepackage{algpseudocode}
\usepackage{svg}

\usepackage{lineno}
\begin{document}
\begin{frontmatter}
\title{Exploiting Semantic Scene Reconstruction for Estimating Building Envelope Characteristics}
\author[1]{Chenghao Xu} 
\author[2]{Malcolm Mielle}
\author[1]{Antoine Laborde}
\author[1]{Ali Waseem}
\author[1]{Florent Forest}
\author[1]{Olga Fink} 

\affiliation[1]{organization={EPFL},
            city={1015 Lausanne},
            country={Switzerland}}
\affiliation[2]{organization={Schindler EPFL Lab},
            city={1015 Lausanne},
            country={Switzerland}}

\begin{abstract}
Achieving the EU's climate neutrality goal requires retrofitting existing buildings to reduce energy use and emissions.
A critical step in this process is the precise assessment of geometric building envelope characteristics to inform retrofitting decisions.
Previous methods for estimating building characteristics, such as window-to-wall ratio, building footprint area, and the location of architectural elements, have primarily relied on applying deep-learning-based detection or segmentation techniques on 2D images.
However, these approaches tend to focus on planar facade properties, limiting their accuracy and comprehensiveness when analyzing complete building envelopes in 3D.

While neural scene representations have shown exceptional performance in indoor scene reconstruction, they remain under-explored for external building envelope analysis. 
This work addresses this gap by leveraging cutting-edge neural surface reconstruction techniques based on signed distance function (SDF) representations for 3D building analysis.
We propose BuildNet3D, a novel framework to estimate geometric building characteristics from 2D image inputs.
By integrating SDF-based representation with semantic modality, BuildNet3D recovers fine-grained 3D geometry and semantics of building envelopes, which are then used to automatically extract building characteristics.
Our framework is evaluated on a range of complex building structures, demonstrating high accuracy and generalizability in estimating window-to-wall ratio and building footprint.
The results underscore the effectiveness of BuildNet3D for practical applications in building analysis and retrofitting.
\end{abstract}



\begin{keyword} Neural Surface Reconstruction \sep Signed Distance Function (SDF) \sep Building Envelope Characteristics \sep Window-to-Wall Ratio



\end{keyword}

\end{frontmatter}



\section{Introduction}
\label{sec:intro}

Building digitalization has revolutionized the design, construction, and operation of assets in the Architecture, Engineering, and Construction (AEC) industry~\cite{watson2011digital, li2023towards}. 
These developments have significantly impacted fields such as automated construction management~\cite{Borrmann2018, EINI2021102222, jiang2021intelligent}, building retrofitting and maintenance~\cite{chen2019bim, motawa2013knowledge}, and energy consumption inspections~\cite{wang-ubem, wei2011design}.
Two important innovations related to building digitalization are: Building Information Modeling (BIM), which provides a digital representation of the physical and functional characteristics of a facility~\cite{Borrmann2018}, and Building Energy Modeling (BEM), which uses physics-based simulations to analyze building energy use~\cite{reinhart2016urban}.
Together, these technologies enhance the ability to design new buildings and retrofit existing structures. Indeed, retrofitting existing building infrastructures and their envelopes is an important task to achieve the climate neutrality target, as buildings in the EU account for 40\% of energy consumption and 35\% of greenhouse gas emissions~\cite{act2011communication}.

The building envelope serves as a physical barrier between a building's interior and the external environment, including components such as roofs, windows, doors, floors, and walls, which covers the entire exterior structure~\cite{sadineni2011passive}.
It plays a critical role in regulating a building's energy efficiency and thermal performance~\cite{nardi2018quantification}.
However, constructing detailed building envelope models for both BIM and BEM remains a time-consuming and labor-intensive task, often requiring manual intervention and professional expertise~\cite{xiong2013automatic, jung2018automated}.
Existing automatic reconstruction methods typically rely on point-cloud data to construct simplified geometric representations of 3D buildings. However, these methods struggle to capture fine-level surface geometry, making the resulting models difficult to update, refine and edit~\cite{Liu2023MeshDiffusion}, especially in the case of complex building structures.

Recent advancements in neural scene representations, such as Neural Radiance Fields (NeRF)~\cite{mildenhall2021nerf} and NeuS~\cite{wang2021neus}, employ Multi-layer Perceptrons (MLP) to learn implicit representations of 3D scenes, significantly improving the performance of 3D reconstruction from sparse sets of RGB images. These cutting-edge techniques enable more accurate and detailed modeling of complex scenes, even with limited input data.
Notably, neural scenes can be integrated with multiple modalities, including depth~\cite{Azinovic_2022_CVPR}, semantics~\cite{Zhi:etal:ICCV2021}, and thermal~\cite{xu2024leveragingthermalmodalityenhance}, to recover corresponding 3D distributions from 2D image inputs.
However, most existing applications focus on interior or large city-scale reconstruction, and their potential for building envelope analysis remains under-explored.

To inform decision-making in building retrofits, precise estimation of building envelope characteristics is essential for assessing current building conditions, especially for existing buildings that lack detailed digital models.
In assessing building energy efficiency, key characteristics such as the window-to-wall ratio\footnote{\url{https://help.buildingenergyscore.com/support/solutions/articles/8000026042-window-to-wall-ratio}} (WWR), building footprint (i.e., the contour of the building at the ground level), and overall dimensions are critical for evaluating energy consumption and heating loads~\cite{ALSHARGABI2022104577, choi2012energy}. 
Current methods for estimating these characteristics primarily rely on 2D images and deep learning techniques to infer these values from single facade captures~\cite{chen2019development, desimone2024windowwallratiodetection}.
However, these approaches often focus on local planar properties of a single facade, overlooking the global geometric relationships between facades and failing to capture valuable three-dimensional information, such as curved walls. This limitation restricts their accuracy and applicability for comprehensive building analysis.

\begin{figure*}[t]
    \centering
    \includegraphics[width=\textwidth]{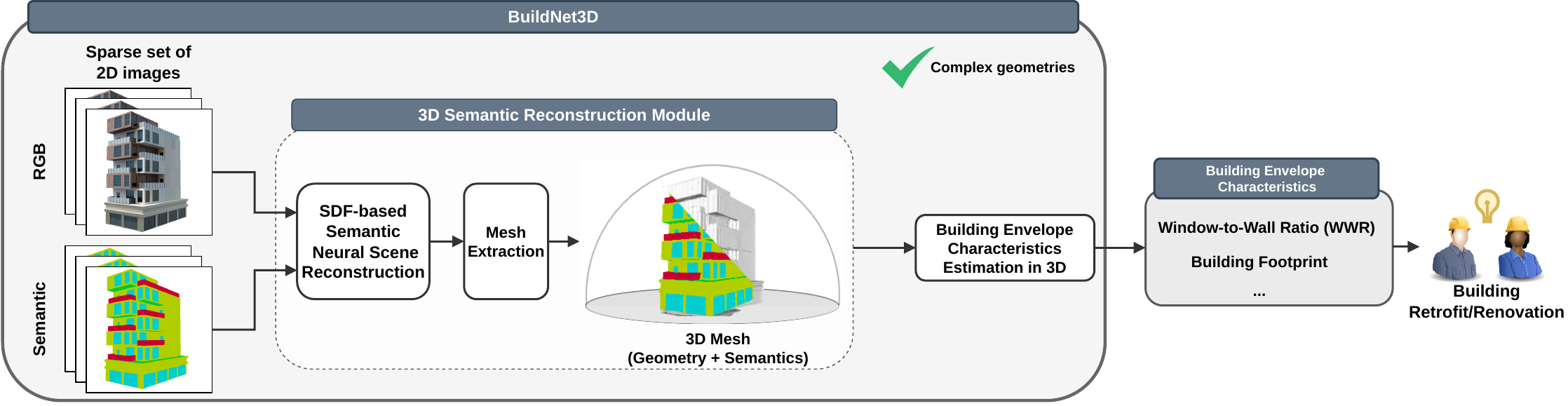}
    \caption{Architecture of the BuildNet3D framework for estimating building characteristics from 2D image inputs. The 2D color images with corresponding semantic images are utilized to reconstruct a semantic building in 3D.}
    \label{fig:pipeline}
\end{figure*}

In this paper, we present BuildNet3D, a novel framework that reconstructs building envelopes and estimates their characteristics in three dimensions from 2D image inputs, leveraging the capabilities of neural scene representations, as illustrated in \cref{fig:pipeline}.
First, we leverage state-of-the-art neural implicit surface representation~\cite{wang2021neus} as our reconstruction module to recover fine-grained geometry and appearance. Additionally, we integrate a semantic network to derive comprehensive 3D semantic information across the building envelope surface.
Second, we introduce an innovative mesh-based approach for estimating building characteristics.
This approach decomposes the reconstructed building mesh into a set of triangular faces, each possessing independent geometric, color, and semantic attributes. 
By synthesizing these properties from the neural scenes, we can accurately estimate corresponding building characteristics from a fully 3D spatial perspective. 

To validate our BuildNet3D framework, we present a novel building dataset accompanied by a dedicated data generation pipeline and ground-truth building characteristics.
This dataset features a variety of 3D building models with increasing geometric complexity and corresponding multi-modal image data. Using these synthetic buildings, we can directly acquire building information and image data from any viewpoint, enabling preliminary validation and benchmarking on our BuildNet3D without the need for real-world data collection and measurements.
We conduct a comprehensive evaluation on various building types within this dataset, focusing on estimating the window-to-wall ratio (WWR) and extracting building footprints.
Furthermore, we propose an improved 2D-semantics-based baseline method for calculating the WWR, which integrates the depth modality to constrain 3D correlations across different facades, providing a comparative benchmark for our framework. 

BuildNet3D achieves an average WWR error within 5\%, and a building footprint accuracy of at mininum 96\%. These results demonstrate the effectiveness of our approach in estimating building characteristics in three dimensions, overcoming  the limitations of planar-based estimation derived from single images. By exploiting the strengths of advanced neural scenes to achieve accurate and detailed 3D envelope reconstructions, BuildNet3D enables more precise and universally applicable building analysis.

Our contributions can be summarized as follows:
\begin{enumerate}
    \item
    We propose BuildNet3D, a novel framework that employs a semantic-based neural SDF method to reconstruct 3D building models. It then accurately estimates geometric building characteristics by leveraging the reconstructed neural scene enriched with semantic information.
    
    \item We introduce a novel open-source building dataset that consists of a variety of 3D building models and corresponding multi-modal image data.
    This dataset includes ground-truth values for building characteristics, enabling effective benchmarking and validation of BuildNet3D.
    
    \item We propose an improved 2D-semantics-based baseline method for calculating the WWR of entire building envelopes. By integrating depth modality, this approach ensures consistent scaling across facades when computing total areas from 2D images, significantly improving the accuracy and reliability of WWR estimations.
    \item We conduct extensive evaluations using our proposed building dataset, comparing BuildNet3D with the proposed 2D-semantics-based baseline. Our framework achieves significantly higher accuracy 
    in both WWR and building footprint estimations. These results demonstrate the generalized performance of BuildNet3D across various building types, including those with complex geometries.
\end{enumerate}

\section{Related Work}
\label{sec:related_work}

\subsection{Building Characteristics Estimation}
Building characteristics, such as window-to-wall ratio (WWR), orientation, dimensions, number of floors, etc., are commonly used as key input parameters for analyzing energy consumption~\cite{ALSHARGABI2022104577, choi2012energy}, designing sustainable buildings~\cite{bank2010integrating, al2020towards, WANG2024112043}, and informing retrofitting strategies~\cite{pichugin2018reliability, asadi2014multi, BUENO2018104}.
However, these applications often assume that these characteristics are known in advance, with limited research dedicated to their estimations from visual inputs.

The WWR is defined as the ratio of the total window or glazing area to the total exterior facade area.
Existing methods for estimating WWR primarily rely on deep-learning-based techniques for window detection on facade images~\cite{NEUHAUSEN2018527, neuhausen2018window, li2020window}. 
After detecting the windows, a post-processing step is applied to the annotated images to calculate the WWR~\cite{desimone2024windowwallratiodetection, shi2020innovative, tueyes} by counting the pixels associated with window and wall classes.
Additionally, \citet{zhuo2023direct} propose an end-to-end regression model that directly computes the WWR from facade images. However, these methods are limited to 2D analysis, neglecting their applicability to 3D building structures, where the estimated WWR is restricted to individual facades rather than capturing the entire building.

In contrast, recent works~\cite{szczesniak2022method, touzani2021machine} seek to calculate the WWR of entire buildings from a 3D spatial perspective. \citet{szczesniak2022method} propose to calculate the WWR by separately counting window pixels on facade views obtained from four predefined orientations, then averaging the WWR values from these views.
However, their method introduces inherent errors due to inconsistent scaling factors between different facades when translating pixel counts into surface areas. \citet{touzani2021machine} reconstruct a simplified 3D building model at Level of Detail 1 (LoD1), which represents buildings as rectangular blocks from drone imagery. They compute the WWR by projecting detected windows onto this 3D model.
Both methods, however,  assume that facades and windows can be modeled as regular planar geometric shapes, overlooking the complexities of real-world 3D structures, such as irregular shapes and  obstructions like staircases, decorative elements.  This simplification highlights a gap in current methodologies, particularly when more detailed spatial analysis is required.

These challenges are not limited to facade and window modeling; they also extend to the accurate delineation of building footprints. Defined as the contour of a building at ground level, accurate building footprints are critical for  applications such as high-definition maps, urban planning, and construction management. These applications  depend on precise measurements of building boundaries to ensure accurate spatial analysis, land use assessment, and infrastructure development, further underscoring the need for more robust techniques that can accommodate real-world structural complexities..

In fact, building footprints can be considered as a combination of various structural  characteristics, such as dimensions, location, and orientation. However, current research primarily focuses on extracting representations of building footprints from single remote sensing images~\cite{guo2021deep, zhu2020map}. 
These methods often face challenges from  occlusions, such as rooftops and  architectural features like balconies and overhangs, which obscure parts of the building. Furthermore, distortions caused by the camera's perspective can complicate the accurate delineation of building footprints from remote sensing data.

As a result, capturing building characteristics for complex structures from a 3D spatial perspective remains a significant challenge. To address this limitation, our work leverages advanced reconstruction techniques that incorporate more accurate and comprehensive 3D geometric information into the estimation of building characteristics.

\subsection{3D Object Reconstruction}
Traditional 3D reconstruction methods for building applications often rely on point-cloud representations. These methods typically use either scan-to-BIM pipelines from laser scans~\cite{gimenez2015review, PEREZ2021108320} or Structure-from-Motion (SfM) techniques from image inputs~\cite{xie2023built,PANTOJAROSERO2022104430}, followed by a post-processing step to extract simplified building models from the point clouds.
However, point-based building models struggle to capture fine-grained building envelope details, making the resulting models unsuitable for subsequent updates, refinements, and simulations~\cite{Liu2023MeshDiffusion}.

Recent advancements in neural scene representation, such as NeRF~\cite{mildenhall2021nerf}, enable novel view synthesis by constructing an implicit 3D scene directly from a set of RGB images. The implicit neural scenes are parameterized by multi-layer perceptrons (MLPs).
One major challenge with implicit representations is extracting precise 3D geometry from neural scenes.
A common approach is to threshold density values to define surfaces~\cite{zeng20163dmatch}. However, this often results in noisy, poorly-defined surfaces due to insufficient constraints. 
To overcome this limitation, recent approaches  convert volume density fields into signed distance function (SDF) representations and introduce an additional eikonal loss to enforce  smooth and consistent surface definitions~\cite{wang2021neus, yariv2021volume, li2023neuralangelo}. In these approaches, the SDF value represents the orthogonal distance from any  point in space to the object's surface, leading to more precise geometry extraction.

Additionally, neural scene representations can be furhter extended by incorporating other modalities, such as surface normals~\cite{Yu2022MonoSDF}, semantic information~\cite{fan2022nerf, Zhi:etal:ICCV2021}, and thermal data~\cite{hassan2024thermonerf, xu2024leveragingthermalmodalityenhance}.
By integrating these multi-modal inputs, neural scene representations can produce more detailed 3D building reconstructions enriched with information like structural semantics, material properties, and thermal performance. 

However, these techniques are currently focused on modeling building geometry and appearance, with limited application in directly extracting key characteristics for building analysis or simulations. Expanding their use to extract meaningful characteristics for practical applications, such as retrofitting guidance, energy efficiency assessments, and construction progress monitoring~\cite{cui20243d}, presents an important research opportunity. This would also facilitate the precise measurement of 3D building dimensions~\cite{jeon2024nerf} and seamless integration into BIM models~\cite{hachisuka2023harbingers}, thereby enhancing the overall utility of neural scene representations for building-related applications.

\section{Preliminary}
\label{sec:preliminary}

\subsection{SDF-Based Volume Rendering}
The implicit scene to be reconstructed is defined by two functions: 
a signed distance function $f(\mathbf{x}): \mathbb{R}^3 \rightarrow \mathbb{R}$, which maps a spatial position $\mathbf{x} \in \mathbb{R}^3$ to its signed distance from the surface, 
and a color function $c(\mathbf{x}, \mathbf{v}) : \mathbb{R}^3 \times \mathbb{S}^2 \rightarrow \mathbb{R}^3$, which maps a point $\mathbf{x} \in \mathbb{R}^3$ and a viewing direction  $\mathbf{v} \in \mathbb{S}^2$ to their corresponding color. 
The object surface $\mathcal{S}$ is represented as the zero-level set of the SDF:
\begin{equation}
\mathcal{S} = \{\mathbf{x} \in \mathbb{R}^{3} | f(\mathbf{x})=0\}
\label{eq:sdf}
\end{equation}

To train the SDF-based scene representation, both the signed distance function $f(\mathbf{x})$ and the color function $c(\mathbf{x}, \mathbf{v})$ are parameterized using Multi-layer Perceptrons (MLP).
The training process is supervised by comparing the synthesized images generated through SDF-based volume rendering~\cite{wang2021neus, yariv2021volume}, with the 2D input images.
Specifically, for each pixel, a ray $\mathbf{r}$ is emitted from the camera origin $\mathbf{o}$ through the pixel along the viewing direction $\mathbf{v}$. Along this ray, $n$ sample points are generated as $\{\mathbf{x}_{i}=\mathbf{o}+t_{i}\mathbf{v} |i=1,\dots,n,t_{i+1} > t_{i} > 0  \}$. The color of the rendered pixel is then computed by accumulating the contributions along the ray as:
\begin{equation}
\hat{C}(\mathbf{r})= \sum_{i=1}^{n} w_{i} c_{i}, \quad \text{where } w_{i}=T_{i} \alpha_{i}
\label{eq:render}
\end{equation}
Here, $c_{i}$ represents the predicted color at each sample point from the color network, $T_{i}=\prod_{j=1}^{i-1}(1-\alpha_{j})$ is the accumulated transmittance, and $\alpha_{i}$ is the opacity at the $i$-th sample point. To convert the volume density predictions from NeRF into SDF representations, the opacity $\alpha_{i}$ is defined as:
\begin{equation}
    \alpha_{i} = \text{max}\left(\frac{\Phi_{s}(f(\mathbf{x}_{i}))-\Phi_{s}(f(\mathbf{x}_{i+1}))}{\Phi_{s}(f(\mathbf{x}_{i}))},0\right)
\end{equation}
where $\Phi_{s}$ is the Sigmoid function with a learnable inverse standard deviation $s$, and $f(\mathbf{x})$ represents the signed distance function. In this work, we use the same SDF-based volume rendering formulation as in NeuS~\cite{wang2021neus}.

\subsection{Neural Surface Reconstruction Loss}
To train the network, the neural SDF and color fields are optimized by minimizing the discrepancy between the rendered color $\hat{C}(\mathbf{r})$ and the ground-truth pixel color ${C}(\mathbf{r})$.
Specifically, the neural networks and the inverse standard deviation $s$ are trained by randomly sampling a batch of pixels and their corresponding camera rays at each iteration. The color loss $\mathcal{L}_\text{color}$ is defined as:
\begin{equation}
\mathcal{L}_\text{color} = \sum_{\mathbf{r} \in \mathcal{R}} \Vert \hat{C}(\mathbf{r}) - C(\mathbf{r}) \Vert_{1}
\end{equation}
where $\Vert \cdot \Vert_{1}$ denotes the $L_1$ loss, and $\mathcal{R}$ is the set of sampled camera rays. 

To regularize the SDF predictions, the eikonal loss $\mathcal{L}_\text{eikonal}$~\cite{icml2020_2086} is introduced, which enforces the SDF gradient to satisfy the eikonal equation $\Vert \nabla f(\mathbf{x}) \Vert_{2}=1$ across all sampled points $\mathcal{X}$:
\begin{equation}
\mathcal{L}_\text{eikonal} = \sum_{\mathbf{x} \in \mathcal{X}} (\Vert \nabla f(\mathbf{x}) \Vert_{2} - 1)^2
\end{equation}

The overall loss function for the neural surface reconstruction is then defined as:
\begin{equation}
\mathcal{L}_\text{surface} = \mathcal{L}_\text{color} + \lambda \mathcal{L}_\text{eikonal}
\end{equation}

\section{Methodology}
\label{sec:method}
The primary objective of this work is to accurately infer 3D building geometry 
and envelope characteristics directly from a set of 2D images. 
To accomplish this, we introduce a two-stage approach that leverages advanced neural scene representations and multi-modal integration for building analysis:
\begin{enumerate}
    \item 3D Reconstruction of Semantic Building Envelopes (\cref{sec:sdf})
    \item Building Characteristics Estimation from the Reconstructed 3D Mesh (\cref{sec:est})
\end{enumerate}

\begin{figure*}[t]
    \centering
    \includegraphics[width=\textwidth]{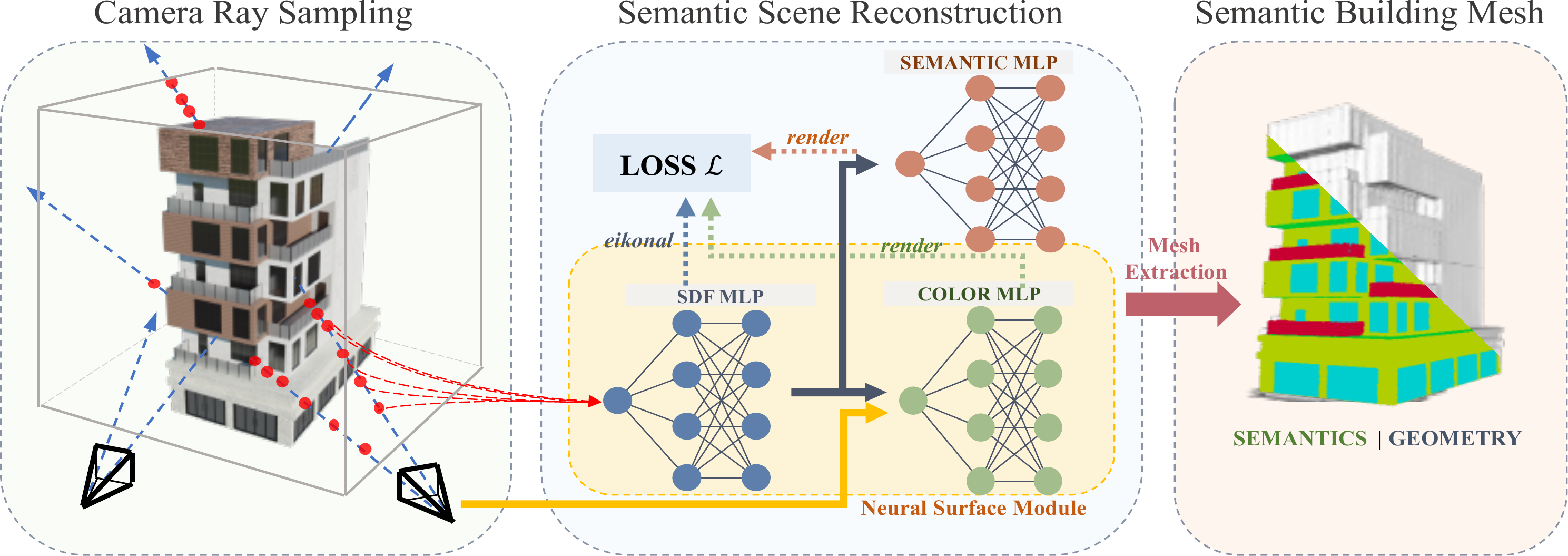}
    \caption{Architecture of reconstruction module for 3D semantic building envelope using SDF-based representation.}
    \label{fig:architecutre}
\end{figure*}

\subsection{Semantic Building Reconstruction}
\label{sec:sdf}
The estimation of building characteristics relies on a 3D mesh that captures both fine-grained geometry and semantic information.
While NeRF-based methods have shown impressive capabilities in novel view synthesis, they often struggle to recover high-fidelity, accurate geometry due to inherent limitations in surface representation through a volume density field.
To address this, we introduce a novel framework that leverages an SDF-based neural surface reconstruction as the backbone, extended with an additional semantic network to predict semantic attributes for spatial points.
This joint optimization framework combines accurate geometric reconstruction with semantic scene segmentation, enabling us to capture detailed building geometry, appearance, and semantics within a 3D scene.
This enriched 3D representation forms the basis for precise estimation of building characteristics.
The architecture of our reconstruction module is illustrated in \cref{fig:architecutre}.

Our proposed semantic building reconstruction process involves the following steps: Given a set of images of the building structure along with corresponding camera poses, we employ the neural surface module described in \cref{sec:preliminary} to reconstruct the building envelope surface $\mathcal{S}$.
To generate the 3D semantic distribution along the surface $\mathcal{S}$, we introduce an additional semantic network that utilizes the encoded features from the SDF network as inputs. A notable advantage of our reconstruction pipeline is its flexibility, as it is agnostic to the specific backbone and can seamlessly integrate with alternative surface reconstruction methods, such as those in \cite{yariv2021volume, li2023neuralangelo, Yu2022MonoSDF}.

The 3D semantic segmentation is modeled as a view-invariant function that maps spatial points $\mathbf{x}$ to semantic logits $\mathbf{s}(\mathbf{x})$, ensuring consistent semantic predictions across different views. Following a formulation similar to image rendering, the semantic logit of a given pixel on the image plane is obtained by:
\begin{equation}
\hat{S}(\mathbf{r})= \sum_{i=1}^{n} T_{i} 
\alpha_{i} \mathbf{s}_{i}
\label{eq:semantic}
\end{equation}
where $\mathbf{s}_{i}$ represents the semantic logit of the sampled point $\mathbf{x}_{i}$ along the camera ray $\mathbf{r}$, and $T_{i}$ and  $\alpha_{i}$ follow the definitions provided in \cref{eq:render}.

The semantic network is trained using ground-truth semantic labels to supervise the learning process. We apply a softmax function to transform the semantic logits into multi-class probabilities $p^{l}(\mathbf{r})$.
The semantic loss is defined as the multi-class cross-entropy loss $\mathcal{L}_{semantic}$ between the rendered semantic output and the ground-truth semantic labels as:
\begin{equation}
\mathcal{L}_\text{semantic} = -\sum_{\mathbf{r} \in \mathcal{R}} \sum_{l = 1}^{L} p^{l}(\mathbf{r}) \log{\hat{p}^{l}(\mathbf{r})}
\end{equation}
where $\mathcal{R}$ is the set of sampled rays within a training batch, $L$ represents the number of semantic classes, and $p^{l}(\mathbf{r})$ and $\hat{p}^{l}(\mathbf{r})$ represent the ground-truth and predicted multi-class semantic probabilities for class $l$ at ray $\mathbf{r}$. 

Hence, the total training loss $\mathcal{L}$ combines the color rendering loss, surface regularization, and semantic segmentation loss as:
\begin{equation}
\mathcal{L} = \mathcal{L}_\text{color} + \lambda_{1} \mathcal{L}_\text{eikonal} + \lambda_{2} \mathcal{L}_\text{semantic}
\end{equation}
where $\lambda_{1}$ and $\lambda_{2}$ are weighting coefficients that control the contribution of the surface regularization and semantic loss, respectively.

\subsection{Building Characteristics Estimation}
\label{sec:est}
After reconstructing the 3D building scenes using our proposed method, we estimate key building envelope characteristics from these models, including the window-to-wall ratio (WWR) and the building footprint.
The WWR estimation involves an explicit mesh extraction process, followed by computing the total areas of both window and facade surfaces. The building footprint is directly extracted from the implicit neural surfaces.

\subsubsection{Window-to-Wall Ratio}
\label{sec:wwr}
Using the surface representation provided by the SDF, we reconstruct the building's surface geometry via the marching cubes algorithm~\cite{marchingcubes}, which extracts a polygonal mesh of the isosurface composed of triangles, with the target isosurface value is set to $0$. 
This explicit mesh representation enables the approximation of complex building geometries through a set of triangle faces, each possessing independent geometric, color, and semantic attributes queried from the neural scenes.

For each 3D triangle face $f_{i} = \{v^{i}_{1},v^{i}_{2}, v^{i}_{3} \}$ in the scene, the corresponding vertex coordinates $\{\mathbf{x}^{i}_{1}, \mathbf{x}^{i}_{2}, \mathbf{x}^{i}_{3} \}$ are retrieved  and passed through the trained network to obtain their semantic labels $\mathbf{s}(\mathbf{x})$.
The areas of window $a_{wd}$ and wall surfaces $a_{wl}$ are then calculated by summing the areas of all $m$ triangles corresponding to the respective semantic categories:
\begin{equation}
    \text{WWR} =  \frac{\sum_{i=1}^{m}(a^{i}_{wd})}{\sum_{i=1}^{m}(a^{i}_{wl})} \label{eq:wwr}
\end{equation}
Since the semantic attributes of each triangle face are derived  from the properties of its vertices, when a triangle has vertices belonging to different semantic classes, its area is weighted according to the proportion of each vertex's class. This is defined as:
\begin{equation}
    a_{wd}= \frac{\text{Num of Window Vertices}}{3} a_t
\end{equation}
\begin{equation}
    a_{wl}= \frac{\text{Num of Wall Vertices}}{3} a_t
\end{equation}
where $a_t$  represents the area of the triangle face.

As discussed in \cref{sec:related_work}, current approaches calculate the WWR by averaging the ratio of window pixels to wall pixels across individual facades using 2D images, without accounting for  the actual dimensions of each facade.
This method ignores scaling differences between facades when aggregating the total window and wall areas for a building. Consequently, the average-based WWR calculation is only accurate when all facades have identical areas or when the scaling between pixel counts and actual areas is consistent across all facades. Moreover, achieving \textbf{ideal perpendicular views} of facades is often impractical in real-world scenarios, particularly for skyscrapers or buildings in dense urban areas where facades may be obstructed by surrounding structures.

\begin{figure*}[t]
    \centering
    \includegraphics[width=\textwidth]{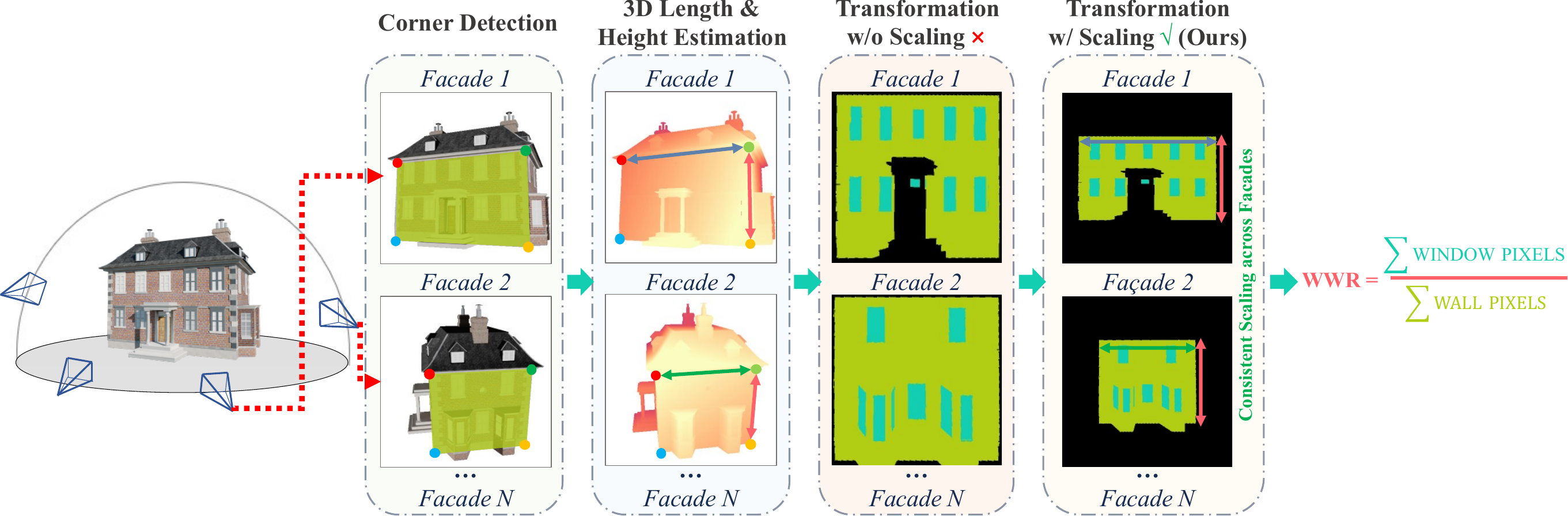}
    \caption{The pipeline of our improved 2D-semantics-based baseline method for estimating WWR of entire building envelopes with consistent scaling factors across all facades.}
    \label{fig:baseline}
\end{figure*}

To ensure a fair baseline comparison, we also propose an improved 2D-semantics-based method that integrates \textbf{multiple natural views} and depth information to maintain consistent scaling during WWR estimation.
Assuming that all $N$ facades of a building share the same height, we calculate the overall building WWR by weighting each facade’s WWR$^i$ according to its length-to-height ratio, $\alpha^i$:

\begin{equation}
    \text{WWR} = \frac{\sum_{i=1}^{N}(s^{i}_{\text{window}})}{\sum_{i=1}^{N}(s^{i}_{\text{facade}})} = \frac{\sum_{i=1}^{N}(\text{WWR}^{i}s^{i}_{\text{facade}})}{\sum_{i=1}^{N}(s^{i}_{\text{facade}})} = \frac{\sum_{i=1}^{N}(\text{WWR}^{i}\alpha^{i})}{\sum_{i=1}^{N}(\alpha^{i})}
\end{equation}
As illustrated in Figure~\ref{fig:baseline}, we extract four corner points from each facade mask to form a quadrilateral that maximally encompasses the facade's area using the provided RGB images and corresponding semantic masks. 
By reprojecting the detected  corner points into 3D space using  the camera's intrinsic parameters and their depth values, we calculate  the actual  length and height of the facade in 3D space from any viewpoint.
To ensure consistent scaling factors between pixel counts and physical facade areas across all images, we apply a homography transformation based on  the facade's length-to-height ratio $\alpha$. This transformation rectifies each facade to an orthogonal (front-facing) view of each building facade, enabling accurate comparison and analysis regardless of the original perspective.

Although the enhanced 2D-semantics method addresses scaling discrepancies between different facades, it still faces limitations in generalizability. First, the input images must contain only a single facade viewed from an angle approximately perpendicular to the surface.
Second, the facade must consist of flat, planar surfaces and cannot accommodate  curved or irregular geometries.
In contrast, our WWR estimation using the BuildNet3D framework can handle arbitrarily complex 3D building geometries without relying on additional depth information.

\subsubsection{Building Footprint}
Accurate extraction of building footprints directly demonstrates BuildNet3D's capability to estimate various building envelope characteristics, such as dimensions, location, and orientation. 
Unlike traditional approaches that derive simplified rectangular representations from aerial images, BuildNet3D focuses on extracting a highly detailed and precise contours of  building boundaries. By capturing the intricate geometries and complex shapes of building envelopes through implicit SDF representations, we provide a more comprehensive and precise analysis of building structures.

To extract precise building footprints, we uniformly sample a dense grid of spatial points at ground level, represented as $\{\mathbf{x_i}=(x_i, y_i, z_i) | z_{i} = 0\}$. Using the trained SDF network, we predict the SDF values for the sampled points and identify the building footprint by detecting  points  that lie near the zero-level set of the SDF. These identified points are then connected to form a closed contour, accurately representing the building footprint.

\section{BuildNet3D Dataset}
\label{sec:dataset}

\begin{table}[ht]
    \centering
    \caption{Building characteristics of synthetic 3D building datasets.}
    \resizebox{1\columnwidth}{!}{
    \begin{tabular}{ccccccc}
    \toprule
     Type & Buildings & Images & Floors & Footprint Range & Height Range & WWR Range\\
     \midrule
     Skyscraper & 12 & 2400 & 9 - 22 & 27.50 - 79.75 & 20.00 - 46.00 & 0.15 - 0.41 \\
     L-shaped & 12 & 2400 & 4 - 6 & 117.00 - 218.00 & 10.00 - 14.00 & 0.16 - 0.32 \\
     Realistic & 4 & 800 & 2 - 7 & 90.29 - 215.04 & 14.46 - 24.86 & 0.10 - 0.20 \\
     \bottomrule
    \end{tabular}}
    \label{tab:dataset}
\end{table}

\begin{figure*}[!ht]
    \centering
    \includegraphics[width=0.95\textwidth]{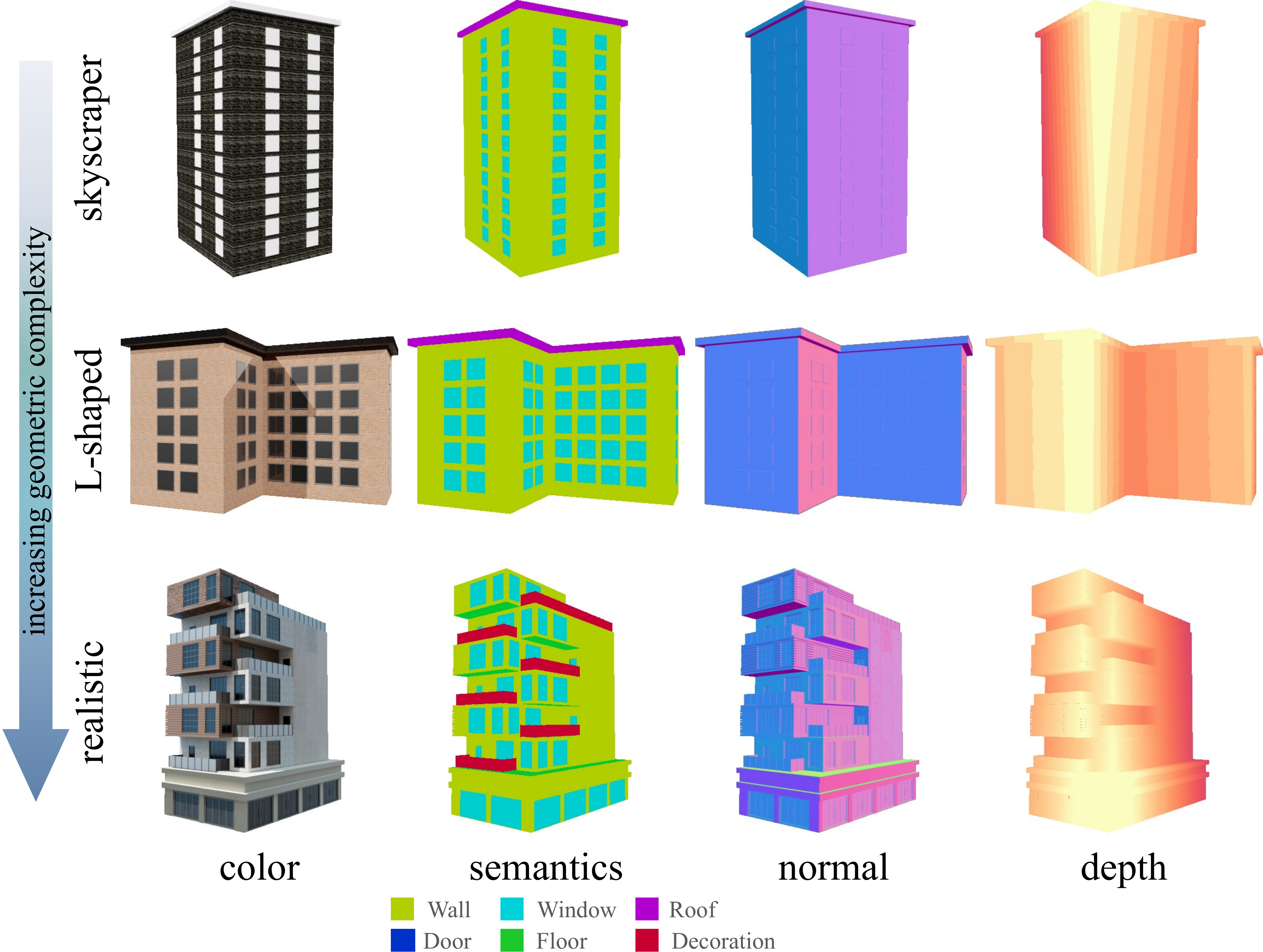}
    \caption{Rendered examples showcasing color, semantic labels, surface normals, and depth images for various structures, including a synthetic skyscraper, a synthetic L-shaped building, and realistic geometrically complex structures, arranged in order of increasing geometric complexity. BuildNet3D utilizes only the \textbf{color} and \textbf{semantic} images. }
    \label{fig:data}
\end{figure*}

To validate our building characteristics estimation framework BuildNet3D, we propose a novel synthetic 3D building benchmark dataset. 
The dataset comprises 28 diverse 3D building models, each accompanied by 200 multi-modal images at $512 \times 512$ pixels resolution and corresponding ground-truth building characteristics.
The building models are categorized into three types, as shown in \cref{fig:data}: \emph{synthetic skyscrapers}, \emph{synthetic L-shaped buildings}, and \emph{realistic geometrically complex structures}. 
Each building type is designed with a range of materials and building characteristics, representing a broad spectrum of architectural complexity. As detailed in \cref{tab:dataset}, the building models exhibit large variations in terms of number of floors, building footprints, building heights, and window-to-wall ratios.

The synthetic skyscraper and L-shaped buildings are generated using the framework provided by \citet{fedorova2021synthetic}. These models simplify building elements, representing walls and windows as basic rectangular shapes and primarily consisting of three key components: walls, windows, and roofs. 
While L-shaped buildings exhibit greater structural complexity compared to skyscrapers, both buildings rely on simplified geometric elements, making them significantly less representative of real-world architecture designs. 
In contrast, the dataset of geometrically complex structures was sourced from freely available 3D models designed by artists and reproductions of real-world buildings. 
These models capture a comprehensive range of building envelope components, including windows, walls, roofs, doors, floors, and decorative features. 
The four realistic buildings include a residential house, a building with rounded facades, and two buildings with extended balconies. 

To render high-fidelity, photo-realistic images of these 3D models, we employ BlenderProc~\cite{Denninger2023}, an advanced modular pipeline built on Blender\footnote{https://www.blender.org/}. Using BlenderProc, we generate  ground truth data, including camera poses, color images, depth maps, normal maps, and semantic labels. The posed color images, paired with their corresponding semantic labels,  serve as inputs for our proposed framework to reconstruct 3D semantic building envelopes. The depth and normal maps are provided to enrich the availability and flexibility of our building datasets, which can potentially be leveraged with other modalities to improve 3D reconstruction, as demonstrated in \cite{Azinovic_2022_CVPR, Yu2022MonoSDF}.

\begin{algorithm}[!ht]
\caption{Camera Pose Sampling}
\label{code:camera}
\begin{algorithmic}[1]
    \State Initialize point of interest (POI) $\mathbf{x}_{\text{POI}}$, set radius $r$
    \State Uniform Sample for $\theta$ and $\phi$ for the spherical coordinate system
    \While{True}
        \State Generate camera location $\mathbf{x}$ based on sampled $\theta$ and $\phi$
        \State Compute camera rotation $\mathbf{R}$ from $\mathbf{x}_{\text{POI}}$ and  $\mathbf{x}$
        \State Generate depth image using ray tracing given camera pose $\{\mathbf{x}, \mathbf{R}\}$
        \State Check boundary conditions of the depth image
    \If{boundary conditions are not met}
        \State Adjust $r$, $\theta$, and $\phi$ to improve boundary conditions
    \EndIf
    
    \If{camera pose is valid \textbf{and} distinct from previous poses}
        \State Store this valid camera pose
        \State \textbf{Break}
    \EndIf
    
    \If{maximum iterations reached \textbf{or} invalid camera pose detected}
        \State Resample $\theta$ and $\phi$
        \State Reset $\mathbf{x}_{\text{POI}}$ and  $r$, restart the process
    \EndIf
\EndWhile
\State Add the valid camera pose to the list of sampled camera poses
\State Repeat until all camera samples are generated
\end{algorithmic}
\end{algorithm}

Each 3D building model is positioned at the center of the scene, with all its elements assigned unique category IDs to facilitate the generation of segmentation labels. 
Additionally, the environment is set with random daylight background images from Polyhaven\footnote{https://polyhaven.com/} to simulate natural lighting conditions, enhancing the realism and quality  of the rendered images.
To capture a diverse set of views, multiple camera poses are sampled around the 3D model.
The virtual cameras are initially placed on a sphere with a specified radius, and the point of interest (POI) is set at the center of the 3D building. The camera viewing direction $\mathbf{v}$ is calculated by  pointing towards the POI using the following equation:
\begin{equation}
    \mathbf{v} = \frac{\mathbf{x}_{\text{poi}}-\mathbf{x}_{\text{camera}}}{||\mathbf{x}_\text{poi}-\mathbf{x}_{\text{camera}||}}
\end{equation}
We employ an iterative method to identify valid camera poses that provide clear, unobstructed views of the scene while avoiding overlap with previously generated positions, as detailed in \cref{code:camera}. 
For each virtual camera, RGB, normal, depth, and segmented views are automatically generated. Additionally, the pipeline can also be extended to provide instance segmentation labels and the precise 3D locations for all building elements.

\section{Experiments}
\label{sec:exp}

In this section, we validate the capabilities of our BuildNet3D framework in estimating the window-to-wall ratio (WWR) and extracting  building footprints.
For WWR estimation, we compare against two baselines: The first baseline model, referred to as \textit{2D-S}, is based on  the method proposed by~\cite{szczesniak2022method} and calculates  the WWR by averaging across all facades. 
The second baseline, referred to as \textit{2D-SS},  represents  our proposed improved 2D semantics method, which incorporates  consistent scaling to enhance WWR estimation.
To more accurately reflect real-world conditions, we evaluate WWR estimation using \textbf{multiple natural views} captured from different angles, rather than \textbf{ideally perpendicular views} of the facade.
We assess all three building types across  three metrics, with the WWR error quantified  as the mean absolute percentage error between the estimated and ground-truth values.
It is important to note that significant errors in estimating window and wall areas can still result in a relatively small WWR error if both areas are misestimated by comparable  factors, thereby maintaining a nearly constant ratio. 
To address this limitation, we separately  assess the area errors, focusing on the discrepancies in the estimated areas of the window and wall classes.

Additionally, we evaluate the accuracy of our building footprint estimation by comparing our estimations with  ground-truth footprints derived from our synthetic building datasets. As there are no existing methods for extracting detailed building footprints, we focus on presenting our estimation results to demonstrate the effectiveness and precision of our approach.

\subsection{Implementation Details}
\noindent\textbf{Network architecture.} 
We use NeuS~\cite{wang2021neus} as the backbone for the neural surface reconstruction module, consisting of an SDF MLP with 8 hidden layers and a color MLP of 4 hidden layers, both with a hidden size of 256, as implemented in nerfstudio~\cite{nerfstudio}.
The SDF network predicts the signed distance $f(\mathbf{x})$, generating intermediate geometric features. The color network takes as input a spatial position $\mathbf{x}$, a viewing direction $\mathbf{v}$, the normal vector of the SDF $\mathbf{n}=\nabla f(\mathbf{x})$, and a 256-dimensional intermediate geometric feature vector, and outputs the corresponding color value $c$. 
The semantic MLP consists of two hidden layers, each containing 128 neurons, which takes the intermediate geometric features as input and outputs a semantic logit vector $\mathbf{s}$.

\noindent\textbf{Training details.}
We configure the size of the scene box to fully encompass the entire building structure,  setting the sphere size parameter during geometry initialization ($\textit{bias}$) to 1.5, and the initial learnable value for transforming SDF to density ($\textit{beta\_init}$) to 0.5. The weighting cofficients for the eikonal loss $\lambda_{1}$ and semantics loss $\lambda_{2}$ are set to $0.1$ and $0.5$, respectively.
The networks are trained using the ADAM optimizer. The learning rate is initially linearly increased from 0 to $5 \times 10^{-4}$ over the first 5,000 iterations, followed  by a cosine decay schedule, gradually reducing it to a minimum of $2.5\times 10^{-5}$. Each neural scene of the building is trained using 100 input images for 100,000 iterations on a single NVIDIA A100 GPU, with a batch size of 2048 rays for all buildings.

\subsection{Window-to-Wall Ratio Estimation}
\begin{table}[t]
    \centering
    \caption{Evaluation of the window-to-wall ratio on the building dataset, including average errors for overall WWR, total window area, and total wall area across four buildings of each type.
    2D-S refers to the baseline 2D-Semantics method, while 2D-SS refers to the improved 2D-Semantics w/ Scaling.
    Ideal-view represents capturing a single perpendicular image per facade, while multi-view involves capturing five images per facade from different positions.}
    \label{tab:wwr}
    \resizebox{0.9\columnwidth}{!}{
    \begin{tabular}{l ccc}
        \toprule
         & WWR Error {[}\%{]} & Window Error {[}\%{]} & Wall Error {[}\%{]} \\
        \midrule
        \multicolumn{4}{c}{Skyscrapers} \\
        \midrule
        2D-S Ideal View & 1.2 $\pm$ 1.4 & 9.4 $\pm$ 5.6 & 10.2$\pm$ 6.9 \\
        2D-S Multi View & 6.0 $\pm$ 2.2 & 14.0 $\pm$ 6.4 & 10.4 $\pm$ 6.9 \\
        2D-SS Ideal View & \textbf{0.6 $\pm$ 0.5} & \textbf{0.8 $\pm$ 0.4} & \textbf{0.2 $\pm$ 0.2} \\
        2D-SS Multi View & 6.7 $\pm$ 1.4 & 7.0 $\pm$ 1.1 & 0.5 $\pm$ 0.2 \\
        BuildNet3D(Ours) & 5.5 $\pm$ 2.0 & 5.2 $\pm$ 1.5 & 0.7 $\pm$ 0.3 \\
        \midrule
        \multicolumn{4}{c}{L-shaped Buildings} \\
        \midrule
        2S-S Ideal View & 10.2 $\pm$ 7.2 & 64.4 $\pm$ 21.2 & 83.0 $\pm$ 17.1 \\
        2S-S Multi View & 8.6 $\pm$ 5.7 & 20.0 $\pm$ 10.8 & 18.1 $\pm$ 16.7 \\
        2S-SS Ideal View & 2.5 $\pm$ 1.3 & 2.2 $\pm$ 1.7 & 1.5 $\pm$ 1.0 \\
         2S-SS Multi View & 8.0 $\pm$ 0.7 & 8.1 $\pm$ 1.4 & 0.6 $\pm$ 0.5 \\
        BuildNet3D(Ours) & \textbf{1.6 $\pm$ 1.3} & \textbf{2.0 $\pm$ 2.0} & \textbf{0.7 $\pm$ 0.4} \\
        \midrule
        \multicolumn{4}{c}{Realistic Buildings} \\
        \midrule
        2D-S Ideal View & 18.7 $\pm$ 14.4 & 37.0 $\pm$ 13.7 & 19.0 $\pm$ 6.8 \\
         2D-S Multi View & 19.5 $\pm$ 16.3 & 37.6 $\pm$ 12.5 & 19.2 $\pm$ 6.0 \\
        2D-SS Ideal View & 10.0 $\pm$ 9.8 & 15.8 $\pm$ 9.8 & 7.0 $\pm$ 3.4 \\
         2D-SS Multi View & 10.4 $\pm$ 12.6 & 15.6 $\pm$ 13.0 & 6.9 $\pm$ 3.3 \\
        BuildNet3D(Ours) & \textbf{3.5 $\pm$ 2.2} &  \textbf{3.3 $\pm$ 2.4} & \textbf{0.8 $\pm$ 0.2} \\
        \bottomrule
    \end{tabular}}
\end{table}

As reported in Tab.~\ref{tab:wwr},  our BuildNet3D demonstrates superior generalizability across all building types. The estimation errors remain stable even as the geometric complexity of building increases, with lower standard deviations observed across four different buildings of the same type. BuildNet3D achieves significantly lower errors—within 5\%—for L-shaped and realistic buildings, and shows a 1\% improvement in performance for skyscrapers compared to the multiple natural view method. Notably, BuildNet3D exhibits relatively lower total wall area errors compared to total window errors, indicating that the extracted triangle mesh effectively approximates the building envelope surface. However, the remaining errors are primarily due to imprecise 3D semantic segmentation when distinguishing between wall and window regions.

As expected, our proposed improved 2D-semantics method (\textit{2D-SS}) effectively reduces area errors by incorporating consistent scaling factors across all facades, showing the best results for simplified skyscrapers. However, its performance deteriorates as geometric complexity increases in L-shaped and realistic buildings, due to its reliance on 2D image inference rather than fully leveraging 3D structural information. Additionally, using natural views introduces more errors due to the challenges of view transformations. Interestingly, in the experiments with L-shaped and realistic buildings, wall area errors are smaller with multiple natural views compared to ideal perpendicular views, as multiple views capture more of the complete facades, which may not be fully visible in a single ideal view. Both BuildNet3D and 2D-SS underscore the importance of incorporating 3D geometric information to improve the accuracy of characteristic estimation across entire building structures.

\subsection{Building Footprint Estimation}
\begin{table}[ht]
    \centering
    \caption{Average IoU of estimated building footprints using BuildNet3D across four buildings of each type.}
    \begin{tabular}{ cccc }
     \toprule
     & Skyscraper & L-shaped & Realistic \\
     \midrule
     IoU [\%] & 0.960 & 0.970 & 0.971\\
     \bottomrule
    \end{tabular}
    \label{tab:footprint}
\end{table}
Building footprint estimation involves not only determining the total area occupied by a building but also accurately localizing and orienting it in 3D space to avoid overlaps with neighboring structures. To assess this, we use the Intersection over Union (IoU) metric between the ground-truth and estimated footprints, where a higher IoU indicates better alignment in terms of area and ground-level positioning:
\begin{equation}
    \text{IoU} = \frac{\text{Area of Intersection}}{\text{Area of Union}}
\end{equation}

As shown in Tab.~\ref{tab:footprint}, all IoU results exceed 96\%, demonstrating impressive accuracy even for geometrically complex buildings and highlighting the generalizability of the method. Minor errors are typically observed at sharp corners, where noise or rounding occurs. These results indicate that our BuildNet3D framework effectively captures and measures building footprints without interference from occlusions.

\section{Conclusion}

In this work, we introduce BuildNet3D, a novel framework that reconstructs 3D building envelope models and estimates key building characteristics from 2D image inputs. By leveraging neural implicit surface representations and integrating a semantic network, BuildNet3D captures fine-grained geometry, appearance, and semantic information, offering a comprehensive 3D geometric understanding for building characteristic estimation.
 
By predicting the window-to-wall ratio and building footprint, BuildNet3D marks a significant advancement in capturing the structural intricacies of building envelopes using SDF representations. It enables precise estimation by utilizing 3D geometric information derived from neural scenes, overcoming the limitations of existing 2D planar-based methods. To validate our approach, we introduced a novel building dataset featuring diverse 3D models with varying geometric complexities. Our evaluations demonstrated that BuildNet3D consistently achieved high accuracy across different building types, with an average WWR error within 5\% and building footprint accuracy exceeding 96\%. These results highlight the robustness and generalizability of BuildNet3D, particularly for complex architectural geometries.

Furthermore, BuildNet3D is a powerful tool for constructing enriched 3D building models, serving as a valuable resource for Building Information Modeling (BIM) and Building Energy Modeling (BEM).
It can be directly used to predict key building characteristics for retrofitting decisions and energy efficiency assessments. 
By integrating diverse data into detailed 3D models, it supports better-informed decisions in architecture, engineering, and urban planning, especially for complex structures and areas with incomplete digital records in real-world scenarios.

In future research, we aim to explore semantic reconstruction with weak supervision, reducing the reliance on extensive  semantic ground truth for all input images. This approach will enhance the scalability and adaptability of the model in real-world scenarios where labeled data is limited. Additionally, we intend to integrate more advanced 3D reconstruction techniques to enhance geometric detail and fidelity, further enriching building envelope models for even more precise building analysis.

\bibliography{biblio}

\end{document}